\documentclass[10pt,twocolumn,letterpaper]{article}

\usepackage{wacv}
\usepackage{times}
\usepackage{epsfig}
\usepackage{graphicx}
\usepackage{amsmath}
\usepackage{amssymb}

\usepackage{multirow}
\usepackage{booktabs}
\usepackage{soul}
\usepackage{amsmath}

\usepackage{float}
\usepackage{subcaption}

\usepackage[table]{xcolor}

\wacvfinalcopy % *** Uncomment this line for the final submission

\def\wacvPaperID{} % *** Enter the wacv Paper ID here

\begin{document}

%%%%%%%%% TITLE
\title{Pay Attention to What You Read:\\ Non-recurrent Handwritten Text-Line Recognition}

% Authors at the same institution
%\author{Lei Kang, Mar\c{c}al Rusi{\~n}ol, Alicia Forn\'{e}s, Pau Riba \\
%Computer Vision Center, Universitat Aut{\`o}noma de Barcelona\\ Barcelona, Spain\\
%{\tt\small \{lkang, marcal, afornes, priba\}@cvc.uab.es}
%\and
% Authors at different institutions
%Mauricio Villegas \\
%omni:us,\\ Berlin, Germany\\
%{\tt\small mauricio@omnius.com}
% \and
% Second Author \\
% Institution2\\
% {\tt\small secondauthor@i2.org}
%}

\author{Lei Kang$^{* \dag}$, Pau Riba$^{*}$, Mar\c{c}al Rusi{\~n}ol$^{*}$, Alicia Forn\'{e}s$^{*}$,  Mauricio Villegas$^{\dag}$ \\
$^{*}$Computer Vision Center, Universitat Aut{\`o}noma de Barcelona, Barcelona, Spain\\
{\tt\small \{lkang, priba, marcal, afornes\}@cvc.uab.es}
\\
$^{\dag}$omni:us, Berlin, Germany\\
{\tt\small \{lei, mauricio\}@omnius.com}
}

%\author{
%    \IEEEauthorblockN{Lei Kang\IEEEauthorrefmark{1}\IEEEauthorrefmark{2}, Mar\c{c}al Rusi{\~n}ol\IEEEauthorrefmark{1}, Alicia Forn\'{e}s\IEEEauthorrefmark{1}, Pau Riba\IEEEauthorrefmark{1}, Mauricio Villegas\IEEEauthorrefmark{2}}\\
%    \IEEEauthorblockA{\IEEEauthorrefmark{1}Computer Vision Center, Universitat Aut{\`o}noma de Barcelona, Barcelona, Spain\\\{lkang, marcal, afornes, priba\}@cvc.uab.es}\\
%    \IEEEauthorblockA{\IEEEauthorrefmark{2}omni:us, Berlin, Germany\\\{lei, mauricio\}@omnius.com}
%}

\maketitle
\ifwacvfinal\thispagestyle{empty}\fi

%%%%%%%%% ABSTRACT
\begin{abstract}
The advent of recurrent neural networks for handwriting recognition marked an important milestone reaching impressive recognition accuracies despite the great variability that we observe across different writing styles. Sequential architectures are a perfect fit to model text lines, not only because of the inherent temporal aspect of text, but also to learn probability distributions over sequences of characters and words. However, using such recurrent paradigms comes at a cost at training stage, since their sequential pipelines prevent parallelization. In this work, we introduce a non-recurrent approach to recognize handwritten text by the use of transformer models. We propose a novel method that bypasses any recurrence. By using multi-head self-attention layers both at the visual and textual stages, we are able to tackle character recognition as well as to learn language-related dependencies of the character sequences to be decoded. Our model is unconstrained to any predefined vocabulary, being able to recognize out-of-vocabulary words, \emph{i.e.} words that do not appear in the training vocabulary. We significantly advance over prior art and demonstrate that satisfactory recognition accuracies are yielded even in few-shot learning scenarios.
%{\color{red} \textbf{TOREWRITE MR:}} Handwritten Text Recognition (HTR) is the problem of transcribing handwritten text image to a sequence of characters, where both the image and text string have variable lengths. The existing state-of-the-art approaches for HTR tasks widely adopt sequence-to-sequence method with Connectionist Temporal Classification (CTC) or Attention mechanism. However, both of the methods are based on Recurrent Neural Network (RNN), whose main limitations are: first, lack of capacity to deal with long handwritten text; second, high computational complexity in non-parallel manner so as to slow down the training speed. To overcome the disadvantages above, we propose, as far as we know, the first non-recurrent system for HTR tasks. We adapt the Transformers with modifications to achieve the state-of-the-art performance on the handwritten data of both word and textline level. Further more, we investigate and discuss the key modules that may have potential to further improve the performance. Finally, we study the effectiveness in low resource cases via transfer learning. 
\end{abstract}

%%
%% The code below is generated by the tool at http://dl.acm.org/ccs.cfm.
%% Please copy and paste the code instead of the example below.
%%

% A "teaser" image appears between the author and affiliation
% information and the body of the document, and typically spans the
% page.
\begin{figure}
  \includegraphics[width=\linewidth]{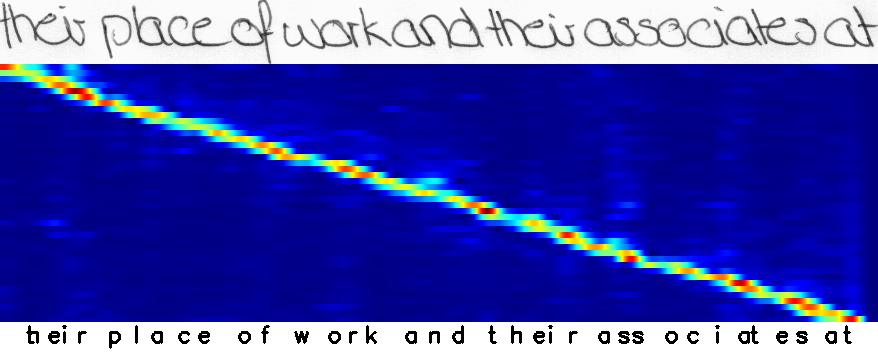}\\
  \hfill\\
  \includegraphics[width=\linewidth]{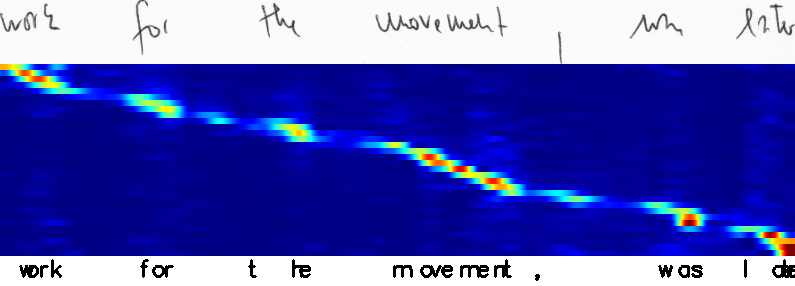}
  \caption{Handwriting text-line recognition with the proposed transformer architecture. The attention mechanism focus at different locations to decode character by character.}
  \label{fig:teaser}
\end{figure}

%%
%% This command processes the author and affiliation and title
%% information and builds the first part of the formatted document.
\maketitle

%\listoftables %% Tro add this shitty thing!! Without this, tables will disappear! Tag: mierda

%%%%%%%%%%%%%%%%%%%%%%%%%%%%%%%%%%%%%%%%%%%%%%%%%%%%%%%%%%%%%%%%%%%%%%
\section{Introduction}

Handwritten Text Recognition (HTR) frameworks aim to provide machines with the ability to read and understand human calligraphy. From the applications perspective, HTR is relevant both to digitize the textual contents from ancient document images in historic archives as well as contemporary administrative documentation such as cheques, forms, etc. Even though research in HTR began in the early sixties~\cite{mermelstein1964system}, it is still considered as an unsolved problem. The main challenge is the huge variability and ambiguity of the strokes composing words encountered across different writers. Fortunately, in most cases, the words to decipher do follow a well defined set of language rules that should be also modelled and taken into account in order to discard gibberish hypotheses and yield higher recognition accuracies. As a result, HTR is often approached by combining technologies from both computer vision and natural language processing communities.

Handwritten text is a sequential signal in nature. Texts are written from left to right in Latin languages, and words are formed by an ordered sequence of characters. Thus, HTR approaches usually adopted temporal pattern recognition techniques to address it. The early approaches based on Hidden Markov Models (HMM)~\cite{bianne2011dynamic} evolved towards the use of Deep Learning techniques, in which Bidirectional Long Short-Term Memory (BLSTM) networks~\cite{graves2008novel} became the standard solution. Recently, inspired by their success in the applications such as automatic translation  or speech-to-text, Sequence-to-Sequence (Seq2Seq) approaches, conformed by  encoder-decoder networks led by attention mechanisms have started to be applied for HTR~\cite{michael2019evaluating}. All the above methods are not only a good fit to process images sequentially, but also have, in principle, the inherent power of language modelling, \emph{i.e.} to learn which character is more probable to be found after another in their respective decoding steps. Nonetheless, this ability of language modelling has proven to be limited, since recognition performances are in most cases still enhanced when using a separate statistical language model as a post-processing step~\cite{tensmeyer2018language}.

Despite the fact that attention-based encoder-decoder architectures have started to be used for HTR with impressive results, one major drawback still remains. In all of those cases, such attention mechanisms are still used in conjunction with a recurrent network, either BLSTMs or Gated Recurrent Unit (GRU) networks. The use of such sequential processing deters parallelization at training stage, and severely affects the effectiveness when processing longer sequence lengths by imposing substantial memory limitations.

Motivated by the above observations, Vaswani \emph{et al.} proposed in~\cite{vaswani2017attention} the seminal work on the Transformer architecture. Transformers rely entirely on attention mechanisms, relinquishing any recurrent designs. Stimulated by such advantage, we propose to address the HTR problem by an architecture inspired on transformers, which dispenses of any recurrent network. By using multi-head self-attention layers both at the visual and textual stages, we aim to tackle both the proper step of character recognition from images, as well as to learn language-related dependencies of the character sequences to be decoded.  

The use of transformers in different language and vision applications have shown higher performances than recurrent networks while having the edge over BLSTMs or GRUs by being more parallelizable and thus involving reduced training times. Our method is, to the best of our knowledge, the first non-recurrent approach for HTR. Moreover, the proposed transformer approach is designed to work at character level, instead at the commonly used word level in translation or speech recognition applications. By using such design we are not restricted to any predefined fixed vocabulary, so we are able to recognize  out-of-vocabulary (OOV) words, \emph{i.e.} never seen during training. Competitive state-of-the-art results on the public IAM dataset are reached even when using a small portion of training data.

The main contributions of our work are summarized as follows. \textbf{\textit{i}}) For the first time, we explore the use of transformers for the HTR task, bypassing any recurrent architecture. We attempt to learn, with a single unified architecture, to recognize character sequences from images as well as to model language, providing context to distinguish between characters or words that might look similar. The proposed architecture works at character level, waiving the use of predefined lexicons. \textbf{\textit{ii}}) By using a pre-training step using synthetic data, the proposed approach is able to yield competitive results with a limited amount of real annotated training data. \textbf{\textit{iii}}) Extensive ablation and comparative experiments are conducted in order to validate the effectiveness of our approach. Our proposed HTR system achieves new state-of-the-art performance on the public IAM dataset.

%%%%%%%%%%%%%%%%%%%%%%%%%%%%%%%%%%%%%%%%%%%%%%%%%%%%%%%%%%%%%%%%%%%%%%
\section{Related Work}

The recognition of handwritten text has been commonly approached by the use of sequential pattern recognition techniques. Text lines are processed along a temporal sequence by learning models that leverage their sequence of internal states as memory cells, in order to be able to tackle variable length input signals. Whether we analyze the former approaches based on HMMs~\cite{bianne2011dynamic,espana2010improving,gimenez2014handwriting} or the architectures based on deep neural networks such as BLSTMs~\cite{graves2008novel},  Multidimensional LSTMs~\cite{graves2009offline,puigcerver2017multidimensional} (MDLSTM) or encoder-decoder networks~\cite{bluche2016joint,kang2018convolve,sueiras2018offline,chowdhury2018efficient,michael2019evaluating}, they all follow the same paradigm. Although all those approaches use recurrent architectures to properly conceal and learn serial information, visually, but also from the language modelling perspective, they all suffer of the lack of parallelization during the training stage. Moreover, in order to efficiently train deep learning based approaches, a huge amount of labeled training data is required. Some approaches like~\cite{bhunia2019handwriting,gurjar2018learning,krishnan2019hwnet} alleviate the cost and effort of collecting such amount of real annotated training data by using synthetically generated cursive data with electronic true-type fonts. Which, in turn, having unlimited annotated data for free and training models that are less prone to overfit to a set of specific writing styles, exaggerate even more the computational costs during the training process.

Vaswani \emph{et al.} presented in~\cite{vaswani2017attention} the Transformer architecture. Their proposal relies entirely on the use of attention mechanisms, avoiding any recurrent steps. Since the original publication, the use of transformers has been popularized in many different computer vision and natural language processing tasks such as automatic translation~\cite{devlin2018bert} or speech-to-text applications~\cite{dong2018speech}. Its use has started to eclipse recurrent architectures such as BLSTMs or GRUs for such tasks, both by being more parallelizable, facilitating training, and by having the ability to learn powerful language modelling rules of the symbol sequences to be decoded.

However, to the best of our knowledge, the transformer architecture has not yet been used to tackle the handwriting recognition problem. It has been nonetheless used lately to recognize text in natural scenes~\cite{lu2019master}. In such works, the original transformers architecture, often applied to one-dimensional signals (\emph{i.e.} text, speech, etc.), has been adapted to tackle two-dimensional input images. Image features are extracted by the use of CNNs~\cite{sheng2019nrtr}, two-dimensional positional encodings~\cite{lee2019recognizing,bleeker2019bidirectional} or additional segmentation modules~\cite{bartz2019kiss} help the system locate textual information amidst background clutter. However, all such works present some limitations when dealing with handwritten text lines. On the one hand, all such architectures work with fixed image size whereas for handwriting recognition we have to face variable length inputs. On the other hand, they work at individual word level, whereas in handwriting recognition we have to face much longer sequences. Finally, despite also having its own great variability, scene text is often much legible than cursive handwriting, since in most of the cases words are formed by individual block letters, which, in turn, are easier to synthesize to obtain large training volumes.

Summarizing, state-of-the-art handwriting recognition based on deep recurrent networks have started to reach decent recognition results, but are too computationally demanding at training stage. Moreover, albeit they shall have the ability to model language-specific dependencies, they usually fall short of inferring adequate language models and need further post-processing steps. In this paper we propose, for the first time, the use of transformers for the HTR task, bypassing any recurrent architecture. A single unified architecture, both recognizes long character sequences from images as well as models language at character level, waiving the use of predefined lexicons.
%%%%%%%%%%%%%%%%%%%%%%%%%%%%%%%%%%%%%%%%%%%%%%%%%%%%%%%%%%%%%%%%%%%%%%
\section{Proposed Method}

\begin{figure*}[t!h]
    \centering
    \includegraphics[width=0.95\linewidth]{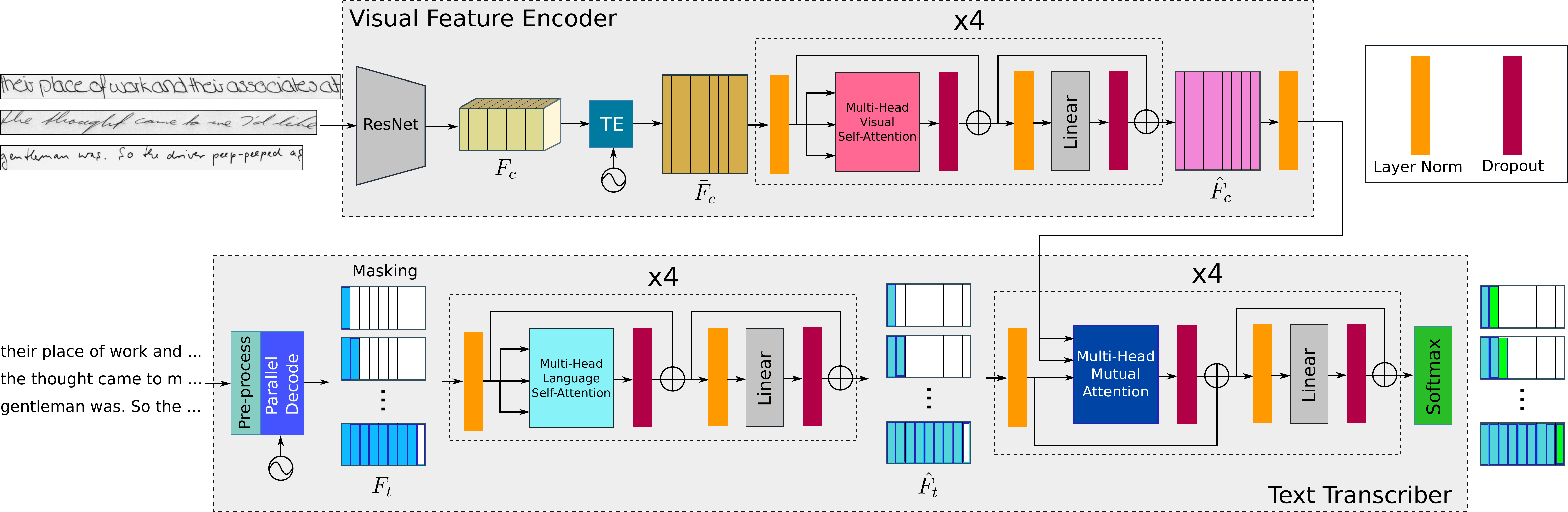}
    \caption{Overview of the architecture of the proposed method. Text-line images are processed by a CNN feature extractor followed by a set of multi-headed self-attention layers. During training, the character-wise embeddings of the transcriptions are also encoded by self-attentions and a final mutual attention module aims to align both sources of information to decode the text-line character by character.}
    \label{fig:arch}
\end{figure*}

%==================================================
\subsection{Problem Formulation}
Let $\{\mathcal{X}, \mathcal{Y}\}$ be a handwritten text dataset, containing images $\mathcal{X}$ of handwritten textlines, and their corresponding transcription strings $\mathcal{Y}$. The alphabet defining all the possible characters of $\mathcal{Y}$ (letters, digits, punctuation signs, white spaces, etc.), is denoted as $\mathcal{A}$. Given pairs of images $x_i \in \mathcal{X}$ and their corresponding strings $y_i \in \mathcal{Y}$, the proposed recognizer has the ability to combine both sources of information, learning both to interpret visual information and to model language-specific rules. 

The proposed method's architecture is shown in Figure~\ref{fig:arch}. It consists of two main parts. On the one hand a visual feature encoder aimed at extracting the relevant features from text-line images and at focusing its attention at the different character locations. Subsequently, the text transcriber is devoted to output the decoded characters by mutually attending both at the visual features as well as the language-related features. The whole system is trained in an end-to-end fashion, learning both to decipher handwritten images as well as modelling language. 

%Our system consists of two main parts: the visual feature encoder and the text transcriber, as shown in Figure~\ref{fig:arch}. The visual feature encoder consists of a cnn feature encoder, a temporal encoding and a visual self-attention module, while the text transcriber consists of a text pre-processing, a character embedding, a language self-attention module and a mutual-attention module. The whole system is trained in an end-to-end fashion, and the details are explained below.

%The alphabet will define the set of textual information that our system is able to transcribe. 
%The alphabet, including all the possible characters (letters, digits, punctuation signs, etc.), is denoted as $\mathcal{A}$.

%==================================================
\subsection{Visual Feature Encoder}
The role of the visual feature encoder is to extract high-level feature representations from an input handwritten image $x\in\mathcal{X}$. It will encode both visual content as well as sequential order information. This module is composed by the following three parts.

\subsubsection{CNN Feature Encoder}
\label{sec:cnn}
Input images $x$ of handwritten text-lines, which might have arbitrary lengths, are first processed by a Convolutional Neural Network. We obtain an intermediate visual feature representation $F_{c}$ of size $f$.  We use the ResNet50~\cite{he2016deep} as our backbone convolutional architecture. Such visual feature representation has a contextualized global view of the whole input image while remaining compact.

\subsubsection{Temporal Encoding}
\label{sec:te}
Handwritten text images are sequential signals in nature, to be read in order from left to right in Latin scripts. The temporal encoding steps are aimed to leverage and encode such important information bypassing any recurrency. 

In a first step, the three-dimensional feature $F_{c}$ is reshaped into a two-dimensional feature by keeping its width, \emph{i.e.} obtaining a feature shape $(f\times h, w)$. This feature map is later fed into a fully connected layer in order to reduce $f\times h$ back to $f$. The obtained feature $F_{c}^{'}$, with the shape of $(f, w)$, can be seen as a $w$-length sequence of visual vectors.

However, we desire that the same character appearing at different positions of the image has different feature representations, so that the attention mechanisms are effectively and unequivocally guided. That is, we want that the visual vectors $F_{c}^{'}$ loose their horizontal shift invariance. Following the proposal from Vaswani \emph{et al.}~\cite{vaswani2017attention}, a one-dimensional positional encoding using sine and cosine functions is applied.

\begin{eqnarray}
    TE(pos, 2i) &=& \sin{\left(\frac{pos}{10000^{2i/f}}\right)}\nonumber\\
    TE(pos, 2i+1) &=& \cos{\left(\frac{pos}{10000^{2i/f}}\right)},
    \label{eq:TE}
\end{eqnarray}
where $pos \in \{0, 1, 2, \ldots, w-1\}$ and $i \in \{0, 1, 2, \ldots, f-1\}$.

$F_{c}^{'}$ and $TE$, sharing the same shape are added along the width axis. A final fully connected layer produces an abscissa-sensitive visual feature $\bar{F}_{c}$ with shape $(f, w)$.

\subsubsection{Visual Self-Attention Module}
\label{sec:sattn}
To further distill the visual features, self-attention modules are applied four times upon $\bar{F}_{c}$. The multi-head attention mechanism from~\cite{vaswani2017attention} is applied using eight heads. This attention module takes three inputs, namely the query $Q_c$, key $K_c$ and value $V_c$, where $Q_c=K_c=V_c=\bar{F}_{c}$. The correlation information is obtained by:

\begin{equation}
    \hat{v}_{c}^{i} = \operatorname{Softmax}\left(\frac{q_{c}^{i}K_{c}}{\sqrt{f}}\right)V_{c},
\end{equation}
where $q_{c}^{i} \in Q_{c}$ and $i \in \{0, 1, \ldots, w-1\}$. The final high-level visual feature is $\hat{F}_{c}=\{\hat{v}_{c}^{0}, \hat{v}_{c}^{1}, \ldots, \hat{v}_{c}^{w-1}\}$.
%==================================================
\subsection{Text Transcriber}
The text transcriber is the second part of the proposed method. It is in charge of outputting the decoded characters, attending to both the visual features as well as the language-specific knowledge learnt form the textual features. It is worth to note that unlike translation of speech-to-text transformer architectures, our text transcriber works at character level instead of word-level. It will thus learn $n$-gram like knowledge from the transcriptions, \emph{i.e.} predicting the next most probable character after a sequence of decoded characters. The text transcriber consists of three steps, the text encoding, the language self-attention step and the mutual-attention module. 

\subsubsection{Text Encoding}
Besides the different characters considered in alphabet $\mathcal{A}$, we require some symbols without textual content for the correct processing of the text-line string. Special character $\langle S\rangle$ denotes the start of the sequence, $\langle E\rangle$  the end of the sequence, and $\langle P\rangle$ is used for padding. The transcriptions $y \in \mathcal{Y}$ are extended to a maximum length of $N$ characters in the prediction. 

A character-level embedding is performed by means of a fully-connected layer that maps each character from the input string to an $f$-dimensional vector. The same temporal encoding introduced in eq.~\ref{eq:TE} is used here to obtain
\begin{equation}
    F_{t} = \operatorname{Embedding}\left(y \right) + TE,
\end{equation}
where $F_{t}$ has the shape of $(f, N)$. 

In the decoding step of recurrent-based HTR approaches~\cite{kang2018convolve,michael2019evaluating} every decoded character is iteratively fed again to the decoder, to predict the next character, thus inhibiting its parallelization. Contrary, in the transformer paradigm, all possible decoding steps are fed concurrently at once with a masking operation~\cite{vaswani2017attention}. To decode the $j$-th character from $y$, all characters at positions greater than $j$ are masked so that the decoding only depends on predictions produced prior to $j$. Such a parallel processing of what used to be different time steps in recurrent approaches drastically reduces training time. 

\subsubsection{Language Self-attention Module}
\label{sec:sattn_t}
This module follows the same architecture as in Section~\ref{sec:sattn} and aims to further distill the text information and learn language-specific properties. $\hat{F}_{t}$ is obtained after the self-attention module implicitly delivers $n$-gram-like features, since to decode the $j$-th character from $y$ all character features prior to $j$ are visible.

%So $F_{t}$ goes through this module and $\hat{F}_{t}$ is obtained sharing the same shape. The language self-attention module shares the same architecture with visual self-attention module but play different roles. Prior to the language self-attention module, there is only one embedding layer to deal with the text as a prior procedure, thus, it is expected to do a harder work than the one at image level. The self-attention module plays a role as an interior language model, which models sequences of coherent characters based on the transcriptions in the training set. So we denote it as language self-attention module. We will evaluate its effectiveness in Section~\ref{sec:abla}.
\begin{figure*}[t!h]
    \centering
    \begin{tabular}{c}
        \includegraphics[height=0.5cm]{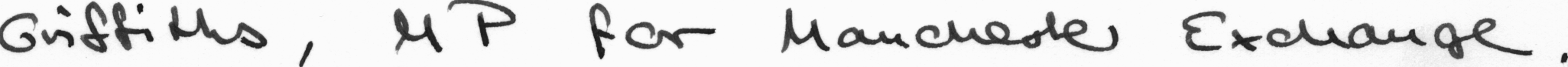}\\
         \includegraphics[height=0.5cm]{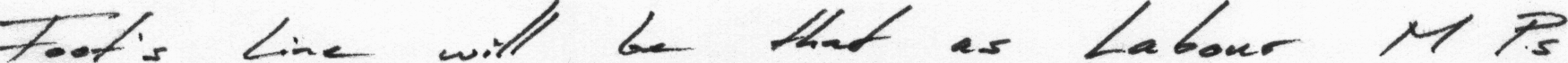}\\
         \includegraphics[height=0.5cm]{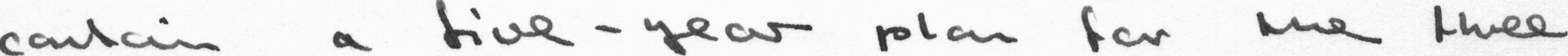}\\
         \includegraphics[height=0.5cm]{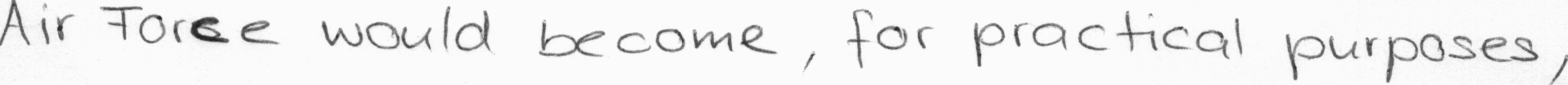}\\
         \includegraphics[height=0.5cm]{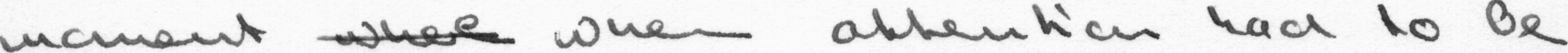}\\
         a) Real data from IAM dataset\\
         \\
         \includegraphics[height=0.8cm]{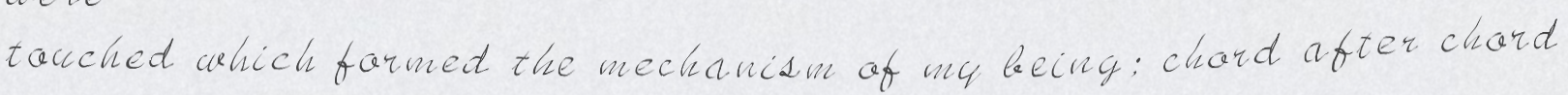}\\
         \includegraphics[height=0.8cm]{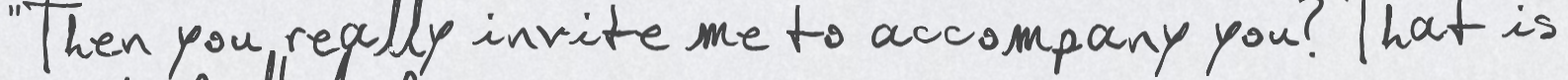}\\
         \includegraphics[height=0.8cm]{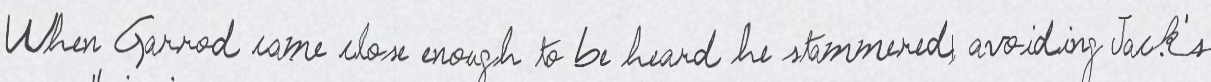}\\
         \includegraphics[height=0.8cm]{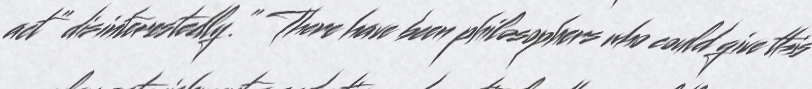}\\
         %\\
         b) Synthetically rendered text-lines with truetype fonts \\
         %\includnegraphics[height=0.8cm]{images/syn/3480-20l.png}\\
         %\includegraphics[height=0.8cm]{images/syn/4225-9l.png}\\
         %\includegraphics[height=0.8cm]{images/syn/806-13l.png}\\
        %  \includegraphics[height=0.5cmw]{images/real/a06-025-04.png}\\
        %  \includegraphics[height=0.5cm]{images/real/a06-064-00.png}\\
        %  \includegraphics[height=0.5cm]{images/real/b06-082-02.png}\\
    \end{tabular}
    \caption{Examples of real and synthetic training handwritten text-line images.}
    \label{fig:syn}
\end{figure*}

\subsubsection{Mutual-attention Module}
\label{sec:mattn}
A final mutual self-attention step is devoted to align and combine the learned features form the images as well as from the text strings. We follow again the same architecture from Section~\ref{sec:sattn}, but now the query $Q_t$ comes from the textual representation $\hat{F}_{t}$ while the key $K_{c}$ and value $V_{c}$ are fed with the visual representations $\hat{F}_{c}$

\begin{equation}
    \hat{v}_{ct}^{i} = \operatorname{Softmax}\left(\frac{q_{t}^{j}K_{c}}{\sqrt{f}}\right)V_{c},
\end{equation}
where $q_{t}^{j} \in Q_{t}$ and $j \in \{0, 1, \ldots, N-1\}$. The final combined representation is  $\hat{F}_{ct}=\{\hat{v}_{ct}^{0}, \hat{v}_{ct}^{1}, \ldots, \hat{v}_{ct}^{N-1}\}$.

The output $\hat{F}_{ct}$ is expected to be aligned with the transcription $Y$. Thus, by feeding the $\hat{F}_{ct}$ into a linear module followed by a softmax activation function, the final prediction is obtained. 
%The visualizations of attention maps can be found in Figure~\ref{fig:attn}, which are done by averaging the mini attention maps across different layers and different heads. Those visualizations could prove the successful alignment after training.

\subsection{Inference on Test Data}
When evaluating on test data, the transcriptions $\mathcal{Y}$ are not available. The text pipeline is initialized by feeding the start indicator $\langle S \rangle$ and it predicts the first character by attending the related visual part on the input handwritten text image. With the strategy of greedy decoding, this first predicted character is fed back to the system, which outputs the second predicted character. This inference process is repeated in a loop until the end of sequence symbol $\langle E\rangle$ is produced or when the maximum output length $N$ is reached. 
%%%%%%%%%%%%%%%%%%%%%%%%%%%%%%%%%%%%%%%%%%%%%%%%%%%%%%%%%%%%%%%%%%%%%%
\section{Experimental Evaluation}
\subsection{Dataset and Performance Measures}
We conduct our experiments on the popular IAM handwritten dataset~\cite{marti2002iam}, composed of modern handwritten English texts. We use the RWTH partition, which consists of $6482$, $976$ and $2914$ lines for training, validation and test, respectively. The size of alphabet $|\mathcal{A}|$ is $83$, including special symbols, and the maximum length of the output character sequence is set to $89$. All the handwritten text images are resized to the same height of $64$ pixels while keeping the aspect ratio, which means that the textline images have variable length. To pack images into mini-batches, we pad all the images to the width of $2227$ pixels with blank pixels.

\emph{Character Error Rate} (CER) and \emph{Word Error Rate} (WER)~\cite{frinken2014continuous} are used for the performance measures. The CER is computed as the Levenshtein distance which is the sum of the character substitutions ($S_c$), insertions ($I_c$) and deletions ($D_c$) that are needed to transform one string into the other, divided by the total number of characters in the groundtruth ($N_c$). Formally, 
    \begin{equation}
	   CER = \frac{S_c + I_c + D_c}{N_c}
	\end{equation}
Similarly, the WER is computed as the sum of the word substitutions ($S_w$), insertions ($I_w$) and deletions ($D_w$) that are required to transform one string into the other, divided by the total number of words in the groundtruth ($N_w$). Formally,
	\begin{equation}
	   WER = \frac{S_w + I_w + D_w}{N_w}
	\end{equation}

\subsection{Implementation Details}

% \begin{figure*}[t!h]
%     \centering
%     \begin{tabular}{c}
%         \includegraphics[height=0.5cm]{images/real/a01-000u-06.png}\\
%          %\includegraphics[height=0.5cm]{images/real/a01-003-03.png}\\
%          %\includegraphics[height=0.5cm]{images/real/a05-000-02.png}\\
%          %\includegraphics[height=0.5cm]{images/real/a05-004-02.png}\\
%          %\includegraphics[height=0.5cm]{images/real/a05-029-05.png}\\
%          %\\
%          %a) Real data from IAM dataset\\
%          %\\
%          %\includegraphics[height=0.5cm]{images/syn/1024-4l.png}\\
%          %\includegraphics[height=0.8cm]{images/syn/1426-13l.png}\\
%          \includegraphics[height=0.5cm]{images/syn/461-21l.png}\\
%          %\includegraphics[height=0.5cm]{images/syn/4707-9l.png}\\
%          %\includegraphics[height=0.5cm]{images/syn/984-11l.png}\\
%          %\\
%          %b) Synthetically rendered text-lines with truetype fonts \\
%          %\includnegraphics[height=0.8cm]{images/syn/3480-20l.png}\\
%          %\includegraphics[height=0.8cm]{images/syn/4225-9l.png}\\
%          %\includegraphics[height=0.8cm]{images/syn/806-13l.png}\\
%         %  \includegraphics[height=0.5cmw]{images/real/a06-025-04.png}\\
%         %  \includegraphics[height=0.5cm]{images/real/a06-064-00.png}\\
%         %  \includegraphics[height=0.5cm]{images/real/b06-082-02.png}\\
%     \end{tabular}
%     \caption{Examples of real, top,  and synthetic, bottom, training handwritten text-line images.}
%     \label{fig:syn}
% \end{figure*}

\subsubsection{Hyper-Parameters of Networks}
In the proposed architecture, the feature size $f$ is $1024$. We use four blocks of visual and language self-attention modules, and each self-attention module has eight heads. We use $0.1$ dropout setting for every dropout layer. In the text transcriber, all the transcriptions include the extended special symbols $\langle S\rangle$ and $\langle E\rangle$ at the beginning and at the end, respectively. Then, they are padded to $89$ length with a special symbol $\langle P \rangle$ to the right, which is the maximum number of characters in the prediction $N$. The output size of the softmax is $83$, which is the size of the alphabet $\mathcal{A}$, including upper/lower cased letters, punctuation marks, blank space and special symbols.

\subsubsection{Optimization Strategy}
We adopt label smoothing mechanism~\cite{szegedy2016rethinking} to prevent the system from making over-confident predictions, which is also a way of regularization. As the ground-truth are one-hot vectors with binary values, label smoothing is done by replacing the $0$ and $1$ with $\dfrac{\varepsilon}{|\mathcal{A}|}$ and $1-\dfrac{|\mathcal{A}|-1}{|\mathcal{A}|}\varepsilon$, where $\varepsilon$ is set to $0.4$ in this paper. We utilize Adam optimizer~\cite{kingma2014adam} for the training process with an initial learning rate of $2 \cdot 10^{-4}$, while reducing the learning rate by half every 20 epochs. The implementation of this system is based on PyTorch~\cite{paszke2017automatic} and performed on a NVIDIA Cluster. The code will be publicly available.

\subsection{Pre-training with Synthetic Data}
\label{sec:syn}
Deep learning based methods need a large amount of labelled training data to obtain a well generalized model. Thus, synthetic data is widely used to compensate the scarcity of training data in the public datasets. There are some popular synthetically generated handwriting datasets available~\cite{krishnan2016generating,kang2019candidate}, but they are at word level. For this reason we have created our own synthetic data at line level for pre-training. First, we collect a text corpus in English from online e-books and end up with over $130,000$ lines of text. Second, we select 387 freely available electronic cursive fonts and use them to randomly render text lines from the first step. Finally, by applying a set of random augmentation techniques (blurring/sharpening, elastic transforming, shearing, rotating, translating, scaling, gamma correcting and blending with synthetic background textures), we obtain a synthetic dataset with $138,000$ lines. The comparison between the synthetic data and the real data is shown in Figure~\ref{fig:syn}.

% \begin{figure*}[t!h]
%     \centering
%     \begin{tabular}{c}
%         \includegraphics[height=0.5cm]{images/real/a01-000u-06.png}\\
%          \includegraphics[height=0.5cm]{images/real/a01-003-03.png}\\
%          \includegraphics[height=0.5cm]{images/real/a05-000-02.png}\\
%          \includegraphics[height=0.5cm]{images/real/a05-004-02.png}\\
%          %\includegraphics[height=0.5cm]{images/real/a05-029-05.png}\\
%          \\
%          a) Real data from IAM dataset\\
%          \\
%          \includegraphics[height=0.5cm]{images/syn/1024-4l.png}\\
%          %\includegraphics[height=0.8cm]{images/syn/1426-13l.png}\\
%          \includegraphics[height=0.5cm]{images/syn/461-21l.png}\\
%          \includegraphics[height=0.5cm]{images/syn/4707-9l.png}\\
%          \includegraphics[height=0.5cm]{images/syn/984-11l.png}\\
%          \\
%          b) Synthetically rendered text-lines with truetype fonts \\
%          %\includnegraphics[height=0.8cm]{images/syn/3480-20l.png}\\
%          %\includegraphics[height=0.8cm]{images/syn/4225-9l.png}\\
%          %\includegraphics[height=0.8cm]{images/syn/806-13l.png}\\
%         %  \includegraphics[height=0.5cmw]{images/real/a06-025-04.png}\\
%         %  \includegraphics[height=0.5cm]{images/real/a06-064-00.png}\\
%         %  \includegraphics[height=0.5cm]{images/real/b06-082-02.png}\\
%     \end{tabular}
%     \caption{Examples of real and synthetic training handwritten text-line images.}
%     \label{fig:syn}
% \end{figure*}

\subsection{Ablation Studies}
\label{sec:abla}
In the ablation studies, all the experiments are trained from scratch with the IAM training set at line-level, and then early-stopped by the CER of the validation set, which is also utilized as an indicator to choose the hyper-parameters as shown in Table~\ref{tab:cnn}~\ref{tab:pe}~\ref{tab:selfattn}.

%Note that instead of a greedy decoding method, the validation measures are done with the input of shifted textline groundtruth, because it is much faster and also what we used in the training process to apply the early stopping. 

\subsubsection{Architecture of CNN Feature Encoder}
We have explored different popular Convolutional Neural Networks for the feature encoder detailed in Section~\ref{sec:cnn}. The best results were obtained with ResNet models. We modified the original ResNet architecture to slighty increase the final resolution of the features, by changing the stride parameter from 2 to 1 in the last convolutional layer. From Table~\ref{tab:cnn}, the best performance is achieved with a modified version of ResNet50.

\begin{table}[t!h]
    \caption{Ablation study on Convolutional architectures. $^*$ indicates modified architectures.}
    \label{tab:cnn}
    \centering
    \begin{tabular}{lcc}
    \toprule
    CNN & CER (\%) & WER (\%)\\
    \midrule
    ResNet34 & 6.33 & 22.63\\
    ResNet34$^*$ & 5.44 & 20.13 \\
    ResNet50 & 5.49 & 20.93 \\
    ResNet50$^*$ & \textbf{4.86} & \textbf{18.65} \\
    \bottomrule
    \end{tabular}
\end{table}

% \begin{figure*}[t!h]
% \centering
% \begin{tabular}{cc}
%     \includegraphics[height=3cm]{images/attn/res_m04-231-01.png}&
%     \includegraphics[height=3cm]{images/attn/res_n04-009-05.png}\\
% \end{tabular}
% \caption{Qualitative results on text-line recognition and visualization of attention maps that coarsely align transcriptions and corresponding image characters.}
% \label{fig:attn}
% \end{figure*}

\subsubsection{Function of Temporal Encoding}
In both the visual feature encoder and the text transcriber, we have used temporal encoding in order to enforce an order information to both visual and textual features. Nonetheless we want to analyze its impact. In Table~\ref{tab:pe}, it is clear that using temporal encoding at text level boosts the performance drastically from $7.72\%$ to $4.86\%$, and from $6.33\%$ to $5.52\%$, depending on whether we use it at image level or not.  The best performance is reached when using the temporal encoding step both for image and text representations.

\begin{table}[t!h]
    \caption{Ablation study on the use of temporal encoding in  image and text levels.}
    \label{tab:pe}
    \centering
    \begin{tabular}{cccc}
    \toprule
    Image level & Text level & CER (\%) & WER (\%)\\
    \midrule
    $-$ & $-$ & 6.33 & 21.64\\
    \checkmark & $-$ & 7.72 & 24.70\\
    $-$ & \checkmark & 5.52 & 20.72\\
    \checkmark & \checkmark & \textbf{4.86} & \textbf{18.65}\\
    \bottomrule
    \end{tabular}
\end{table}

\begin{table*}[t!h]
    \caption{Fine-tuning with different portions of real data (line-level test set with greedy decoding).}
    \label{tab:fine}
    \centering
    %\small
    \begin{tabular}{lcccccccccccccc}
    \toprule
      & \multicolumn{2}{c}{\textbf{20\%}} && \multicolumn{2}{c}{\textbf{40\%}} && \multicolumn{2}{c}{\textbf{60\%}} && \multicolumn{2}{c}{\textbf{80\%}}&& \multicolumn{2}{c}{\textbf{100\%}}\\
    \cmidrule{2-3}\cmidrule{5-6}\cmidrule{8-9}\cmidrule{11-12}\cmidrule{14-15}
    & CER & WER && CER & WER && CER & WER && CER & WER && CER & WER\\
    \midrule
    Seq2Seq  & 20.61 & 56.50 && 16.15 & 46.97 && 15.61 & 46.01 && 12.18 & 38.11 && 11.91 & 37.39\\
    + Synth & 18.64 & 51.77 && 13.01 & 39.72 && 13.00 & 39.34  && 12.15 & 37.43 && 10.64 & 33.64 \\
    \midrule
    Ours & 73.81 & 132.74 && 17.34 & 42.57 && 10.14 & 30.34 && 10.11 & 29.90 && 7.62 & 24.54\\
    \textbf{+ Synth} & \textbf{6.51} & \textbf{20.53} && \textbf{6.20} & \textbf{19.69} && \textbf{5.54} & \textbf{17.71} && \textbf{4.90} & \textbf{16.44} && \textbf{4.67} & \textbf{15.45}\\
    \bottomrule
    \end{tabular}
\end{table*}

\subsubsection{Role of Self-Attention Modules}
Self-attention modules have been applied in both image and text levels. In Table~\ref{tab:selfattn} we analyze their effect in our system. We observe that the visual self-attention module barely improves the performance. Nonetheless, for the language self-attention module, it really plays an important role that improves the performance from $7.71\%$ to $4.86\%$, and from $7.78\%$ to $4.89\%$, with and without the visual self-attention module, respectively. Our intuition is that the language self-attention module actually does learn language-modelling information. This implicitly learned language model is at character level and takes advantage of the contextual information of the whole text-line, which not only boosts the recognition performance but also keep the capability to predict out-of-vocabulary (OOV) words.

\begin{table}[t!h]
    \caption{Ablation study on visual and language self-attention modules.}
    \label{tab:selfattn}
    \centering
    \begin{tabular}{cccc}
    \toprule
    Image level & Text level & CER (\%) & WER (\%)\\
    \midrule
    $-$ & $-$ & 7.78 & 29.78\\
    \checkmark & $-$ & 7.71 & 28.50\\
    $-$ & \checkmark & 4.89 & 18.57\\
    \checkmark & \checkmark & \textbf{4.86} & \textbf{18.65}\\
    \bottomrule
    \end{tabular}
\end{table}

\begin{figure*}[t!h]
    \centering
    \begin{tabular}{cc}
    \includegraphics[width=.45\textwidth]{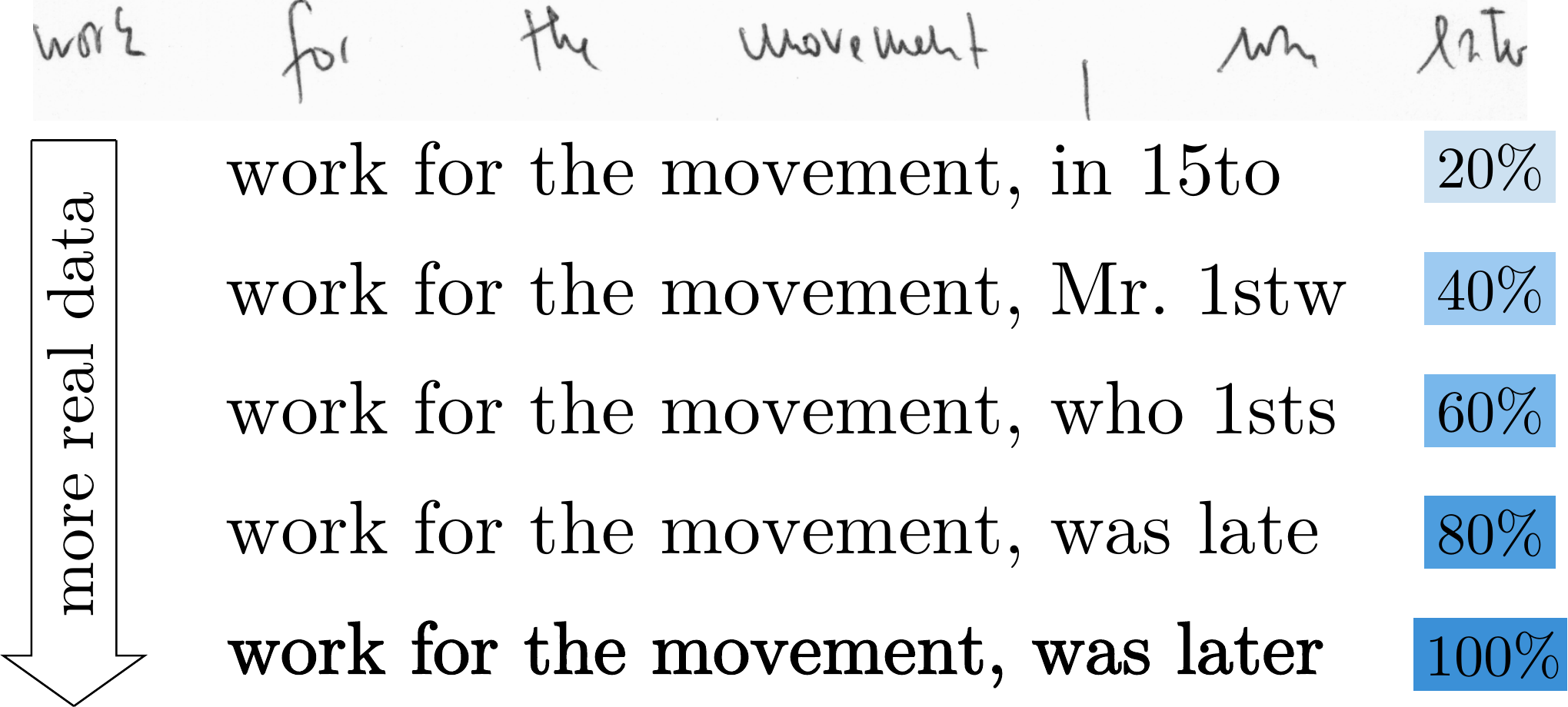}&
    \includegraphics[width=.45\textwidth]{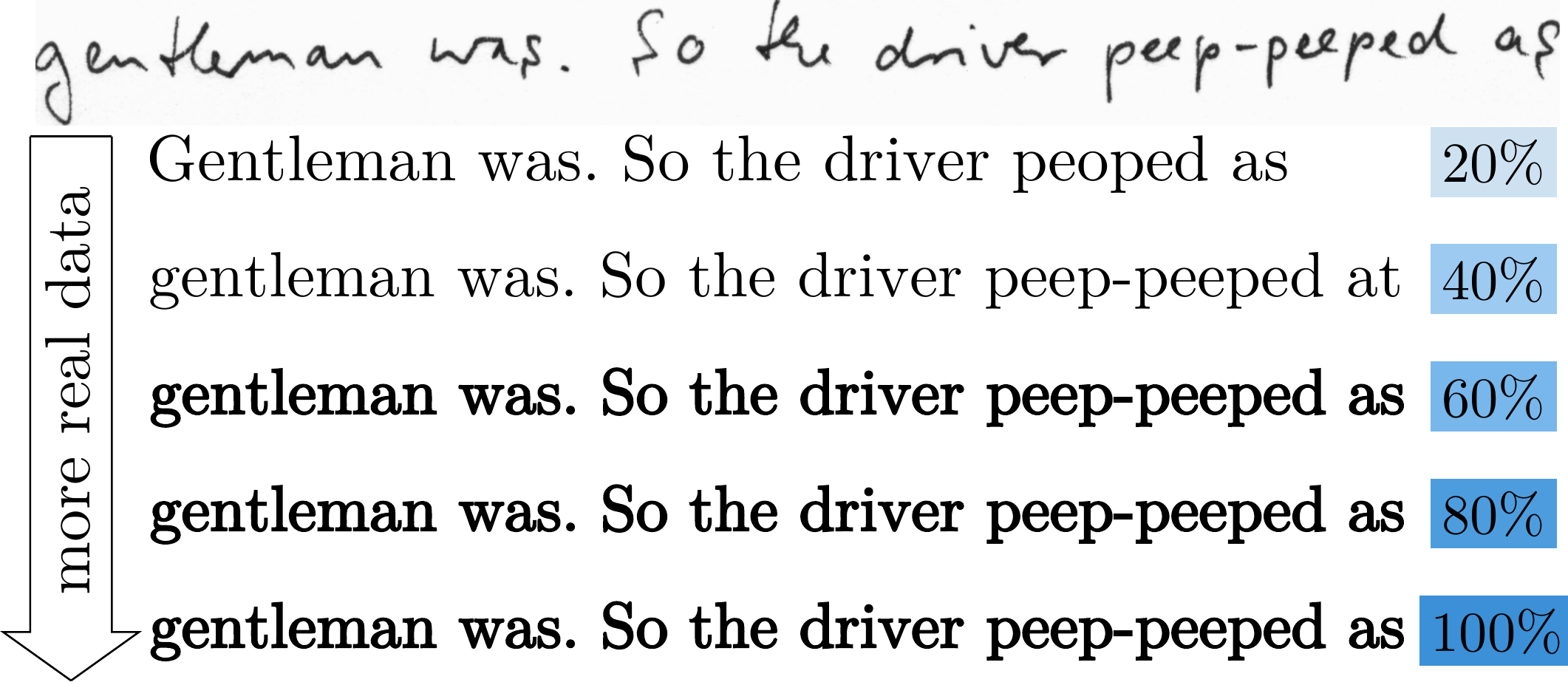}\\
    \includegraphics[width=.45\textwidth]{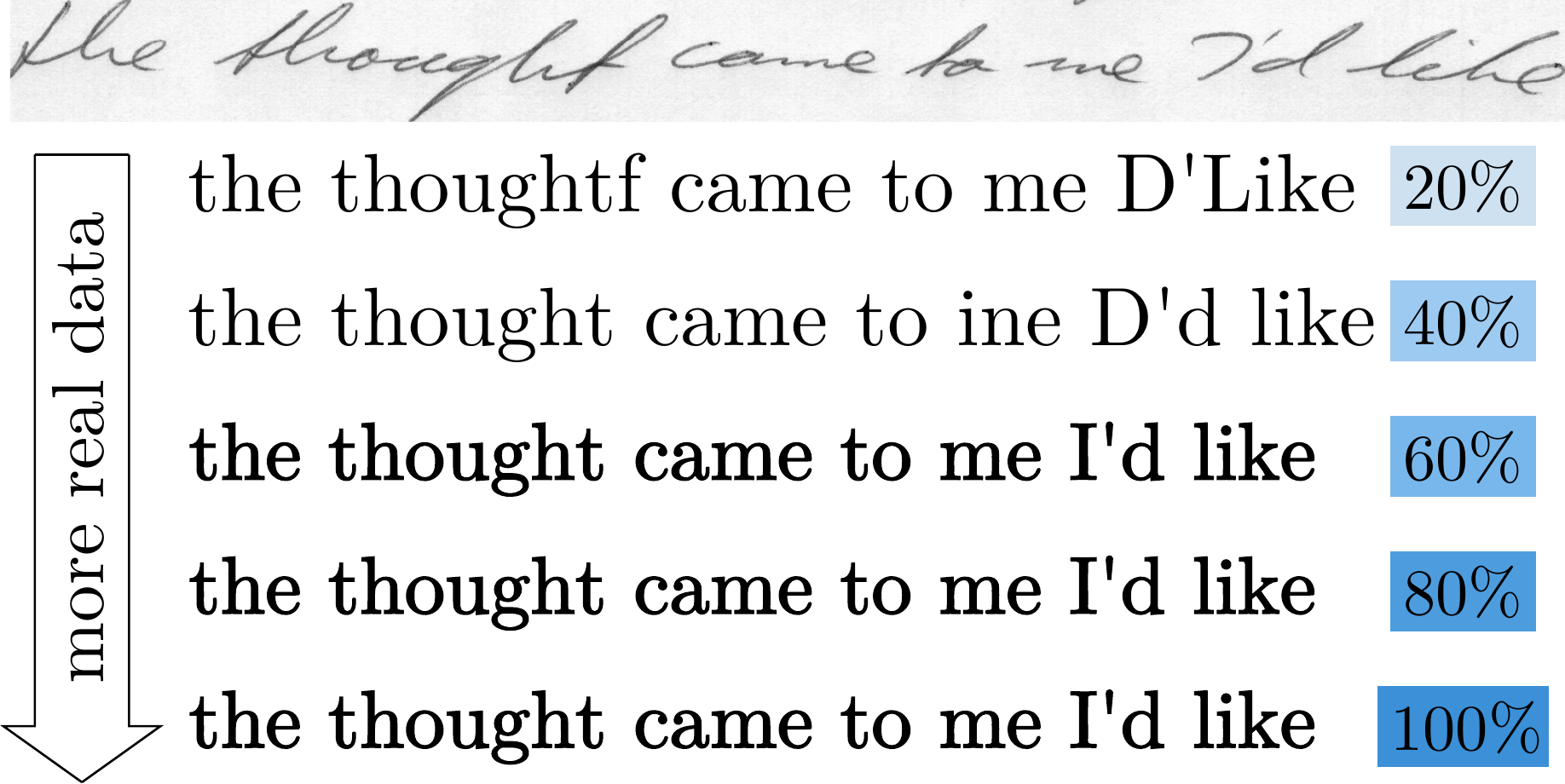}&
    \includegraphics[width=.45\textwidth]{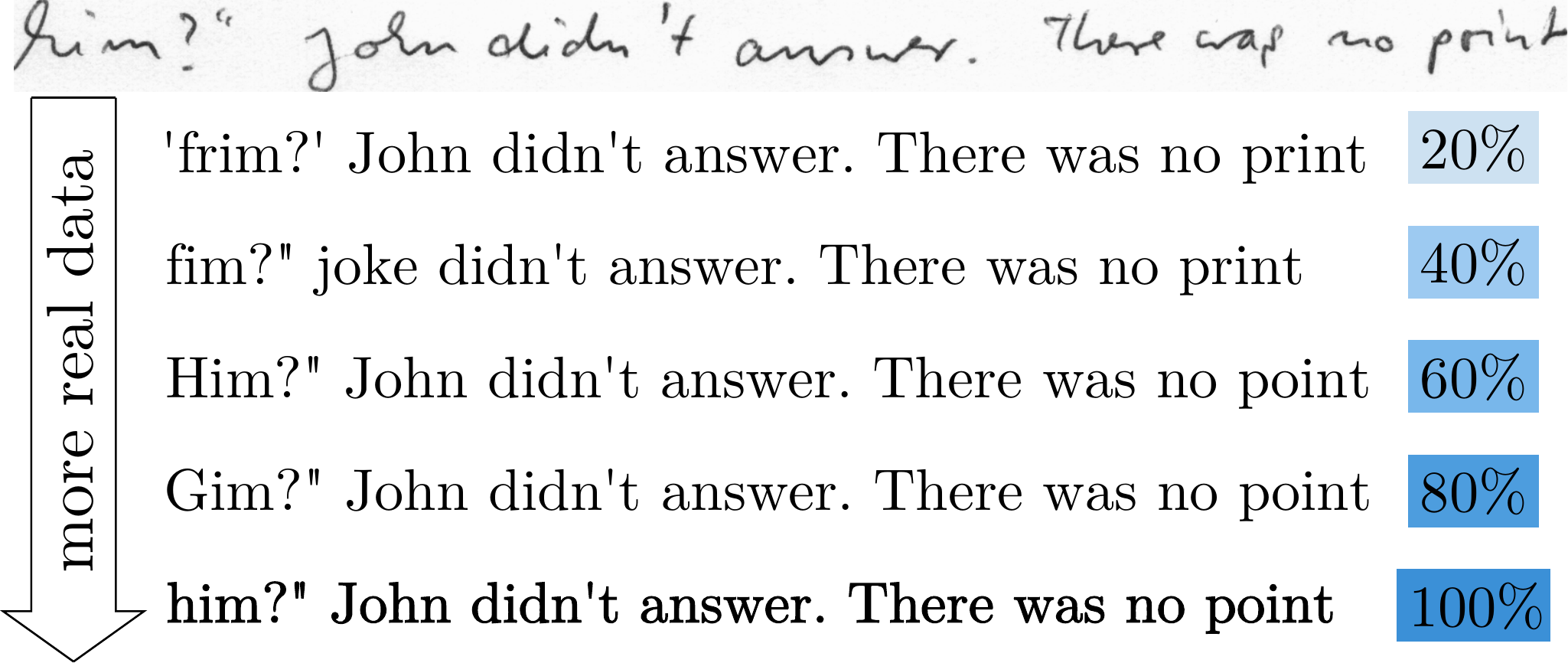}\\
    \end{tabular}
    \caption{Performance of the transformer-based decodings for different amounts of real training data.}
    \label{fig:after_title}
\end{figure*}
% \subsection{Qualitative Results}
% \begin{figure*}[t!h]
% \centering
% \begin{tabular}{cc}
%     \includegraphics[width=0.45\linewidth]{images/attn/res_m04-231-01.png}&
%     \includegraphics[width=0.45\linewidth]{images/attn/res_n04-009-05.png}\\
% \end{tabular}
% \caption{Qualitative results on text-line recognition and visualization of attention maps that coarsely align transcriptions and corresponding image characters.}
% \label{fig:attn}
% \end{figure*}

\begin{figure}[t!h]
\centering
\begin{tabular}{c}
    \includegraphics[width=0.9\linewidth]{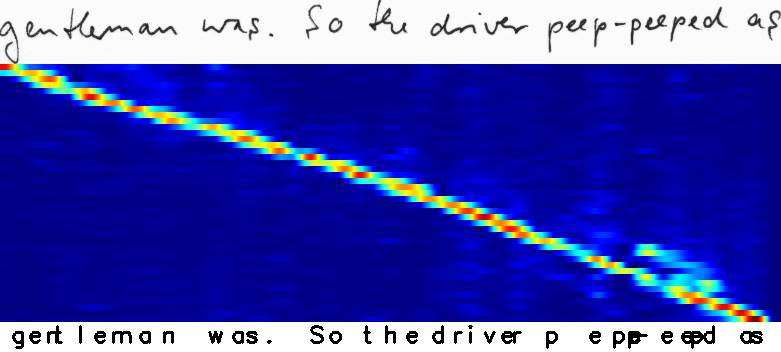}\\
    \includegraphics[width=0.9\linewidth]{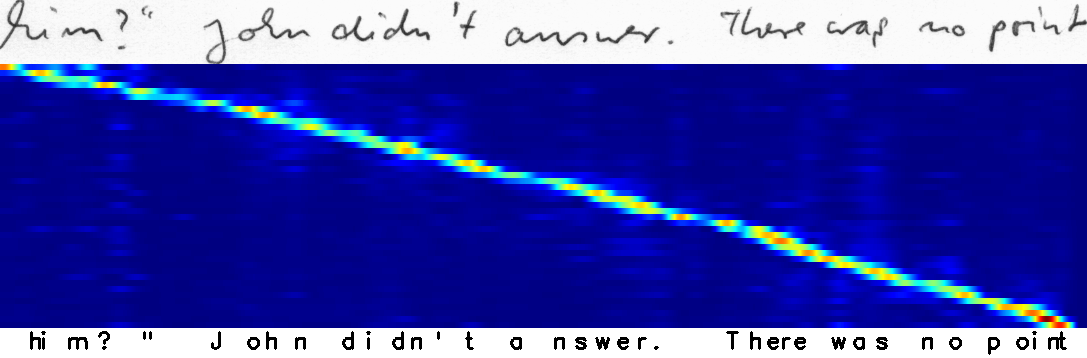}\\
\end{tabular}
\caption{Qualitative results on text-line recognition and visualization of attention maps that coarsely align transcriptions and corresponding image characters.}
\label{fig:attn}
\end{figure}

We showcase in Figure~\ref{fig:teaser} and Figure~\ref{fig:attn} some qualitative results on text-line recognition, where we visualize the attention maps as well. The attention maps are obtained by averaging the mini attention maps across different layers and different heads. Those visualizations prove the successful alignment between decoded characters and images.

\subsection{Detailed Comparison with Seq2Seq Model}
\label{sec:adv_s2s}
In order to provide a fair comparison between the proposed architecture and recurrent-based solutions, we re-implemented a state-of-the-art recurrent handwriting recognition pipeline, and we train and evaluate those under the exact same circumstances. Following the methods proposed in~\cite{kang2018convolve,michael2019evaluating} we built a sequence-to-sequence recognizer composed of an encoder, a decoder and an attention mechanism. The encoder consists of a VGG19-BN~\cite{simonyan2014very} and a two-layer Bidirectional Gated Recurrent Units (BGRU) with feature size of $512$. The decoder is a two-layer one directional GRU with feature size of $512$, and we power the architecture with a location-based attention mechanism~\cite{chorowski2015attention}. All the dropout layers are set to $0.5$. Label smoothing technique is also used during the training process. The maximum number of predicted characters is also set to $89$. All the hyper-parameters in this sequence-to-sequence model are also exhaustively validated by ablation studies with validation data.

We first provide in Table~\ref{tab:comp}, the CER and WER rates on the IAM test set both when training the networks from scratch and just using the IAM training data, and when pre-training the networks with synthetic data for a later fine-tuning step on real data. We also provide the model size and the time taken per epoch during training. While the sequence-to-sequence model has much less parameters, it still takes longer to train than the transformers-based one. We also observe that both models benefit from the use of synthetic pre-training, improving the final error rates quite noticeably for the transformers model, although such boost is not so drastic for the sequence-to-sequence approach. 

\begin{table}[t!h]
    \caption{Comparison between Recurrent and Transformers.}
    \label{tab:comp}
    \centering
    \begin{tabular}{lcccc}
    \toprule
    Method & CER (\%) & WER (\%) & Time(s) & Param(M)\\
    \midrule
    Seq2Seq & 11.91 & 37.39 & 338.7 & 37\\
    + Synth & 10.64 & 33.64 & 338.7 & 37\\
    \midrule
    Ours & 7.62 & 24.54 & 202.5 & 100\\
    \textbf{+ Synth} & \textbf{4.67} & \textbf{15.45} & \textbf{202.5} & \textbf{100}\\
    \bottomrule
    \end{tabular}
\end{table}

% \begin{table*}[t!h]
%     \caption{Fine-tuning with different portions of real data (line-level test set with greedy decoding).}
%     \label{tab:fine}
%     \centering
%     %\small
%     \begin{tabular}{lcccccccccccccc}
%     \toprule
%       & \multicolumn{2}{c}{\textbf{20\%}} && \multicolumn{2}{c}{\textbf{40\%}} && \multicolumn{2}{c}{\textbf{60\%}} && \multicolumn{2}{c}{\textbf{80\%}}&& \multicolumn{2}{c}{\textbf{100\%}}\\
%     \cmidrule{2-3}\cmidrule{5-6}\cmidrule{8-9}\cmidrule{11-12}\cmidrule{14-15}
%     & CER & WER && CER & WER && CER & WER && CER & WER && CER & WER\\
%     \midrule
%     Seq2Seq  & 20.61 & 56.50 && 16.15 & 46.97 && 15.61 & 46.01 && 12.18 & 38.11 && 11.91 & 37.39\\
%     + Synth & 18.64 & 51.77 && 13.01 & 39.72 && 13.00 & 39.34  && 12.15 & 37.43 && 10.64 & 33.64 \\
%     \midrule
%     Ours & 73.81 & 132.74 && 17.34 & 42.57 && 10.14 & 30.34 && 10.11 & 29.90 && 7.62 & 24.54\\
%     \textbf{+ Synth} & \textbf{6.51} & \textbf{20.53} && \textbf{6.20} & \textbf{19.69} && \textbf{5.54} & \textbf{17.71} && \textbf{4.90} & \textbf{16.44} && \textbf{4.67} & \textbf{15.45}\\
%     \bottomrule
%     \end{tabular}
% \end{table*}

\subsection{Few-shot Training}
Due to the scarcity and the cost of producing large volumes of real annotated data, we provide an analysis on the performance of the proposed approach when dealing with a few-shot training setup, when compared again with the sequence-to-sequence approach. To mimic a real scenario in which only a small portion of real data is available, we randomly selected $20\%$, $40\%$, $60\%$ and $80\%$ of the IAM training set.

As shown in Table~\ref{tab:fine}, both sequence-to-sequence and transformer-based approaches follow the same trend. The more real training data is available, the better the performance is. Overall, the transformer-based method performs better than the sequence-to-sequence, except for the extreme case of just having a $20\%$ of real annotated training data available. The transfomer approach, being a much larger model, struggles at such drastic data scarcity conditions. However, when considering the models that have been pre-trained with synthetic data, the transformer-based approach excels in few-shot setting conditions. We provide in Figure~\ref{fig:after_title} some qualitative examples of the transcriptions provided by different models trained with reduced training sets. All of the models were pre-trained with synthetic data.

\subsection{Language Modelling Abilities}
In order to validate whether the proposed approach indeed is able to model language-specific knowledge besides its ability to decode handwritten characters, we propose to test whether using a state-of-the-art language model as a post-processing step actually improves the performance. We implement a shallow fusion~\cite{gulcehre2015using} language model, consisting of a recurrent network with $2,400$ LSTM units. It has been trained on $130,000$ English text-lines. The additive weight for the shallow fusion is set to $0.2$.

We observe in Table~\ref{tab:lm}, that the use of such language modelling post-processing is useless, somehow indicating that the proposed approach already incorporates such language-specific contextual information within the language self-attention module.

\begin{table}[ht!]
    \caption{Effect of using a post-processing langauge model.}
    \label{tab:lm}
    \centering
    \begin{tabular}{lcc}
    \toprule
    Method & CER (\%) & WER (\%)\\
    \midrule
    Ours & 4.67 & \textbf{15.45}\\
    +LM & \textbf{4.66} & 15.47\\
    \bottomrule
    \end{tabular}
\end{table}

\begin{table*}[ht!]
    \caption{Comparison with the State-Of-The-Art approaches on IAM line level dataset.}
    \label{tab:soa}
    \centering
    %\small
    \begin{tabular}{clccc}
    \toprule
    System & Method & $\Omega$ (k) & CER (\%) & WER (\%)\\
    \midrule
    \multirow{10}{*}{
        \begin{tabular}{l}
         HMM/ANN\\
         2008 - now
    \end{tabular}}
    & Almaz\'an~\emph{et al.}~\cite{almazan2014word} & $-$ & 11.27 & 20.01\\
    & Espa\~na~\emph{et al.}~\cite{espana2010improving} & $-$ & 9.80 & 22.40\\
    \cmidrule{2-5}
    
    & Dreuw~\emph{et al.}~\cite{dreuw2011hierarchical} & 50 & 12.40 & 32.90\\
    & Bertolami~\emph{et al.}~\cite{bertolami2008hidden} & 20 & $-$ & 32.83\\
    & Dreuw~\emph{et al.}~\cite{dreuw2011confidence} & 50 & 10.30 & 29.20\\
    & Zamora~\emph{et al.}~\cite{zamora2014neural} & 103 & 7.60 & 16.10\\
    & Pastor~\emph{et al.}~\cite{pastor2015combined} & 103 & 7.50 & 19.00\\
    & Espa\~na~\emph{et al.}~\cite{espana2010improving} & 5 & 6.90 & 15.50\\
    & Kozielski~\emph{et al.}~\cite{doetsch2013improvements} & 50 & 5.10 & 13.30\\
    & Doetsch~\emph{et al.}~\cite{doetsch2014fast} & 50 & 4.70 & 12.20\\
    \midrule
    \multirow{10}{*}{
        \begin{tabular}{l}
         RNN+CTC\\
         2008 - now
    \end{tabular}}
    & Chen~\emph{et al.}~\cite{chen2017simultaneous} & $-$ & 11.15 & 34.55\\
    & Pham~\emph{et al.}~\cite{pham2014dropout} & $-$ & 10.80 & 35.10\\
    & Krishnan\emph{et al.}~\cite{krishnan2018word} & $-$ & 9.78 & 32.89\\
    & Wigington~\emph{et al.}~\cite{wigington2018start} & $-$ & 6.40 & 23.20\\
    & Puigcerver~\cite{puigcerver2017multidimensional} & $-$ & 5.80 & 18.40\\
    & Dutta~\emph{et al.}~\cite{dutta2018improving} & $-$ & 5.70 & 17.82\\
    \cmidrule{2-5}
    & Graves~\emph{et al.}~\cite{graves2008novel} & 20 & 18.20 & 25.90\\
    & Pham~\emph{et al.}~\cite{pham2014dropout} & 50 & 5.10 & 13.60\\
    & Puigcerver~\cite{puigcerver2017multidimensional} & 50 & 4.40 & 12.20\\
    & \textbf{Bluche~\emph{et al.}~\cite{bluche2017gated}} & \textbf{50} & \textbf{3.20} & \textbf{10.50}\\
    \midrule
    \multirow{3}{*}{
        \begin{tabular}{l}
         Seq2Seq\\
         2016 - now
    \end{tabular}}
    & Chowdhury~\cite{chowdhury2018efficient} & $-$ & 8.10 & 16.70\\
    & Bluche~\cite{bluche2016joint} & $-$ & 7.90 & 24.60\\
    \cmidrule{2-5}
    & Bluche~\cite{bluche2016joint} & 50 & 5.50 & 16.40\\
    \midrule
    \textbf{Transf.} & \textbf{Ours} & \textbf{$-$} & \textbf{4.67} & \textbf{15.45}\\
    \bottomrule
    \end{tabular}
\end{table*}

\subsection{Comparison with the State-Of-The-Art}
Finally, we provide in Table~\ref{tab:soa} and extensive performance comparison with the state of the art. Different approaches have been grouped into a taxonomy depending on whether they are based on HMMs or early neural network architectures, whether they use recurrent neural networks (usually different flavours of LSTMs) followed by a Connectionist Temporal Classification (CTC) layer, or if they are based on encoder-decoder sequence-to-sequence architectures. Within each group, we differentiate results depending on whether they make use of a closed vocabulary of size $\Omega$ or they are able to decode OOV words. Bluche~\emph{et al.}~\cite{bluche2017gated} achieves the best result among the methods using a closed lexicon, while our proposed method obtains the best result among the methods without using a closed lexicon, while still competing with most of the closed-vocabulary approaches.

%%%%%%%%%%%%%%%%%%%%%%%%%%%%%%%%%%%%%%%%%%%%%%%%%%%%%%%%%%%%%%%%%%%%%%
\section{Conclusion}
In this paper, we have proposed a novel non-recurrent and open-vocabulary method for handwritten text-line recognition. As far as we know, it is the first approach that adopts the transformer networks for the HTR task. We have performed a detailed analysis and evaluation on each module, demonstrating the suitability of the proposed approach. Indeed, the presented results prove that our method not only achieves the state-of-the-art performance, but also has the capability to deal with few-shot training scenarios, which further extends its applicability to real industrial use cases. Finally, since the proposed approach is designed to work at character level, we are not constrained to any closed-vocabulary setting, and transformers shine at combining visual and language-specific learned knowledge.

{\small
\bibliographystyle{ieee}
\bibliography{bib}
}

\end{document}